\documentclass{article}

\usepackage{arxiv}

\usepackage[utf8]{inputenc} 
\usepackage[T1]{fontenc}    
\usepackage{hyperref}       
\usepackage{url}            
\usepackage{booktabs}       
\usepackage{amsfonts}       
\usepackage{nicefrac}       
\usepackage{microtype}      
\usepackage{lipsum}		
\usepackage{graphicx}
\usepackage{natbib}
\usepackage{doi}

\usepackage{subfig}
\usepackage{subfloat}
\usepackage{booktabs} 
\usepackage{adjustbox}
\usepackage{caption}
\usepackage{hyperref}
\usepackage{multirow}
\usepackage{rotating}
\usepackage{amsmath}
\usepackage{amsfonts}
\usepackage{amssymb}
\usepackage{wrapfig}
\usepackage{authblk}
\usepackage{booktabs, makecell}

\newtheorem{thm}{Theorem}
\newtheorem{col}{Corollary}

\newtheorem{definition}{Definition}

\title{Layer-wise Adaptive Graph Convolution Networks Using Generalized Pagerank}

\author[1]{Kishan Wimalawarne}
\author[1,2]{Taiji Suzuki}
\affil[1]{Department of Mathematical Informatics,  The University of Tokyo, Tokyo, Japan}
\affil[2]{Center for Advanced Intelligence Project (AIP), RIKEN, Tokyo, Japan}
\date{}

\hypersetup{
pdftitle={Layer-wise Adaptive Graph Convolution Networks Using Generalized Pagerank},
pdfsubject={},
pdfauthor={Kishan Wimalawarne, Taiji Suzuki},
}

\begin{document}
\maketitle

\begin{abstract}
 We investigate adaptive layer-wise graph convolution in deep GCN models. We propose AdaGPR to learn generalized Pageranks at each layer of a GCNII network to induce adaptive convolution. We show that the generalization bound for AdaGPR is bounded by a polynomial of the eigenvalue spectrum of the normalized adjacency matrix in the order of the number of generalized Pagerank coefficients.  By analysing the generalization bounds we show that  oversmoothing depends on both the  convolutions by the higher orders of the normalized adjacency matrix and the depth of the model.  We performed evaluations on node-classification using benchmark real data and show that AdaGPR provides improved accuracies compared to existing graph convolution networks while demonstrating robustness against oversmoothing. Further, we demonstrate that analysis of coefficients of layer-wise generalized Pageranks allows us to qualitatively understand  convolution at each layer enabling model interpretations.   
\end{abstract}


\section{Introduction}
In recent years Graph Convolution Networks (GCN) have gained increased recognition as a versatile tool to learn from graphs. Graph convolution networks use the graph topological structures among the data to extract nonlinear features to perform learning tasks. Many recent advances in graph convolution networks have  produced state of the art performances in applications such as social influence prediction \citep{li-goldwasser-2019-encoding}, relationship modelling \citep{10.1007/978-3-319-93417-4_38}, and recommendation systems \citep{3219819.3219890} 

Despite the promising capabilities  and many novel approaches, GCN still faces several limitations that hinders its  full potential in learning with graphs. A well known limitation with GCN is oversmoothing \citep{DBLP:conf/iclr/OonoS20}, where stacking of multiple  convolution layers leads to drop in performance. Oversmoothing is prominent with a model like the Vanila GCN \citep{Kipf:2016tc}, since multiple convolutions by global graph data lead to generalized features that lack the ability learn from labelled data. 
Recently, many approaches have been proposed to mitigate the effect of oversmoothing. Some of these methods  include simple data processing such as  data normalization by Pair-Norms \citep{DBLP:conf/iclr/ZhaoA20} and random removal of edges  using dropedges \citep{DBLP:conf/iclr/RongHXH20}. Many other methods use more complex methods such as random walks as employed in ScatterGCN \citep{DBLP:journals/corr/abs-2003-08414} and skipping layers  as with JKNet \citep{pmlr-v80-xu18c}.
A notable recent development is GCNII \citep{pmlr-v119-chen20v}, which uses  scaled residual layers  and addition of the initial layer  to each convolution layer. GCNII has reported strong robustness against oversmoothing, however, it often requires a deep network to gain a considerable high accuracy.

Another limitation that we identify with GCN is the lack of adaptability of  graph convolution at each layer. Most GCN models apply the same graph convolution method to each layer of a deep network \citep{Kipf:2016tc,pmlr-v119-chen20v}. This not only cause oversmoothing but it may also lead to redundant memory usages and computations.
Furthermore, most GCN models do not provide a systematic approach to understand and interpret   graph convolutions applied at each layer of a deep model.
In practice, to design a optimal GCN model it is  desirable to know the suitable  graph convolution method to apply as well as the amount of convolution to be applied at each layer depending on the data and the learning task.  Recently proposed  GPR-GNN \citep{chien2021adaptive}  learns a generalized Pagerank within  the APPNP model \citep{KlicperaBG19} to perform adaptive graph convolution. However,   GPR-GNN is a shallow network and  does not consider graph convolution in multiple layers, hence, it may not be efficient as a  deep GCN model. 

In this paper, we investigate adaptive convolution in deep graph convolution networks. In contrast to the widely adapted view of applying the same graph convolution method at each layer, we propose that graph convolution should be different for each layer. In our view, graph convolution should be adaptive in a layer-wise manner where the GCN model should be able to learn how to apply graph convolution depending on the network architecture, nature of the data, and the learning task. We propose AdaGPR to apply adaptive generalized Pageranks at each layer of a GCNII model by learning to predict the coefficients of generalized Pageranks using sparse solvers. 
We also give a new generalization error analysis of AdaGPR in which the Rademacher complexity is given as a polynomial of the eigenvalue spectrum of the normalized adjacency matrix. 
This bound reflects the mixing information effect, more specifically, the oversmoothing effect,
and thus yields a better generalization error bound for a graph with a large node degree. 
We conduct evaluations on node-classification and show that  AdaGPR provides better accuracy compared to state  of the art GCN methods.  As a further advantage of our method, we demonstrate that analysis of the coefficients of layer-wise generalized Pagerank  allows us to quantitatively understand layer-wise convolution leading to semi-interpretable GCN models. 

\section{Review}
We start by defining notations used in this paper. Let $G = (V,E)$ a graph with nodes $v_i \in V, \; i=1,\ldots,N$ and edges $(v_i,v_j) \in E$. Let $X \in \mathbb{R}^{N \times q}$ represents a feature matrix with each row representing $q$ features. Let $Y \in \mathbb{R}^{N \times c}$ represents labels of the $N$ nodes with each consisting of $c$ classes. The adjacency matrix of $G$ is represented as $A \in \mathbb{R}^{N \times N}$, and the self-loops added adjacency matrix is  $\hat{A} = A + I_{N}$, where  $I_N \in \mathbb{R}^{N \times N}$ is a identity matrix. We denote the diagonal degree matrix of $\hat{A}$ by $\hat{D}_{ij} = \sum_{k} \hat{A}_{ik}\delta_{ij}$, then the normalized adjacency matrix is $\tilde{A} = \hat{D}^{-1/2}\hat{A}\hat{D}^{-1/2}$.  

The most simple graph convolution network (also known as the Vanilla GCN) was proposed in~\citet{Kipf:2016tc}, where each layer of a multilayer network is multiplied by the normalized graph adjacency matrix before applying a nonlinear activation function. A $2$-layer Vanilla GCN is given as
\begin{equation*}
Z = \mathrm{softmax}\big(\tilde{A} \mathrm{ReLU}(\tilde{A}XW_{0})W_{1}\big), \label{eq:gcn}
\end{equation*}
where $W_0 \in \mathbb{R}^{q \times h}$ and $W_1 \in \mathbb{R}^{h \times c}$ are learning weights with $h$ hidden units. It is well observed that the Vanilla GCN model is highly susceptible to oversmoothing with the increase of depth \citep{DBLP:conf/iclr/OonoS20,pmlr-v119-chen20v}. 

Recently,  many methods that have been proposed to overcome oversmoothing \citep{DBLP:conf/iclr/ZhaoA20,pmlr-v119-chen20v}. One of the successful methods robust agasint oversmoothing with the increase of convolution layers is GCNII \citep{pmlr-v119-chen20v}. It  proposes to multiply convolution at each layer and add the initial layers with sclaing. 
The resulting $l+1$th convolution layer  of GCNII is given as
\begin{equation}
    H^{(l+1)} = \sigma\Big( \big((1-\alpha_l)\tilde{A}H^{(l)} + \alpha_l H^{(0)} \big)  \big((1-\beta_l) I_{N} + \beta_l W^{(l)}\big) \Big), \label{eq:gcnII}
\end{equation}
where  $\sigma(\cdot)$ is the ReLU operator, $H^{(0)} = \sigma(XW^{(0)})$ is the output from initial layer, $W^{(0)}$ and $W^{(l)}$ are weight matrices, and $\alpha_l \in [0,1]$ and $\beta_l \in [0,1]$ are user-defined parameters.

Another approach that resaerhers have adaptoed to overcome oversmoothness in graph convolution models is to use the personalized Pagerank \citep{BRIN1998107} instead of the convolution by the adjacency matrix. PPNP and APPNP \citep{KlicperaBG19} are tow methods that uses the personalized Pagerank convolution to obtain improved accuracy for node classification. A computationally feasible methods that avoid matrix inversion of the personalized Pagerank is the generalized Pagerank (GPR) \citep{e74bd842bfa94260a66fd477178c58e3}, which is defined with $K$ powers of the normalized adjacency matrix with  coefficients  $\mathbf{\mu} = [\mu_{0},\ldots,\mu_{K-1}] \in [0,1]^{K}$ as  
\begin{equation}
\mathrm{GPR}(\mathbf{\mu}) := \sum_{k=0}^{K-1} \mu_{k}\tilde{A}^{k}. \label{eq:GPR}  
\end{equation}
The advantage of using GPR is the ability to learn the coefficients  $\mathbf{\mu}$ from the data \citep{e74bd842bfa94260a66fd477178c58e3}. Adaptive learning of GPR is used by GPRGNN \citep{chien2021adaptive} given by the following model, 
\begin{equation}
P = \mathrm{softmax}(Z),\; Z = \sum_{k=0}^{K-1} \mu_{k} H^{(k)}, \;   H^{(k)} = \tilde{A}H^{(k-1)}, \;   H_{i:}^{(0)} = f_{\theta}(X_{i:}), \label{eq: GPR-GNN} 
\end{equation} 
where $\theta$ represents learning parameters of a multilayer network and $\mu$ is learned using message passing.

There are several limitations in above models. Both personalized Pagerank based models and  GPR-GNN  apply only a single convolution by variants of the Pagerank on the learned representation  prior to the output layer.  Further, these models do not apply any learning weights and nonlinear activation functions after convolution, hence, they do not create  deep GCN models. On the other hand, the GCNII allows us to develop deep models that are robust against the oversmoothing, however, it relies solely on convolutions by the normalized adjacency matrix lacking any adaptive convolution  or any benefits offered by the Pagerank.  

\section{Proposed Method}
We propose adaptive layer-wise graph convolution for deep graph conventional models. Our approach is simple, where we propose to apply a generalised Pagerank at each layer of the GCNII and learn  coefficients of  generalised Pageranks.   

As in GCNII, we use a initial layer $H^{(0)} = \sigma(XW^{(0)})$ without any graph convolution using learning weights $W^{(0)} \in \mathbb{R}^{N \times h}$, where $h$ is the number of hidden units. Given $L$ layers of graph convolutions,  we replace the convolution by $\tilde{A}$ at layer $l$ of \eqref{eq:gcnII} with the generalized Pagerank \eqref{eq:GPR} using $K$ orders  of $\tilde{A}$ and  coefficients $\mathbf{\mu}^{(l)} = (\mu_{0}^{(l)},\ldots,\mu_{K-1}^{(l)}) \in [0,1]^{K}$. Additionally, we impose the constraint $\sum_{k=0}^{K-1} \mu_{k}^{(l)} =1$.  In order to make generalized Pagerank adaptive for each layer, the model needs to learn coefficients $\mu^{(l)} \in \mathbb{R}^{K}, \; l=1,\ldots,L$ by using separate  learning weights $v^{(l)} \in \mathbb{R}^{K},\;l=1,\ldots,L$, respectively. Furthermore, we provide  flexibility to apply a suitable activation function $g(\cdot)$ on $v^{(l)}$  in order to obtain specific properties such as sparseness. We call the new graph convolution network AdaGPR, where its $(l+1)$th layer is defined as 
\begin{equation}
    H^{(l+1)} = \sigma\Bigg( \bigg((1-\alpha_l) \bigg(\sum_{k=0}^{K-1} \mu^{(l)}_k \tilde{A}^{k} \bigg) H^{(l)} + \alpha_l H^{(0)} \bigg)\big((1-\beta_l) I_{N} + \beta_l W^{(l)}\big) \Bigg), \;\; \mu^{(l)} = g(v^{(l)}) \label{eq:AdaGPR},
\end{equation}
where $W^{(l)} \in \mathbb{R}^{h \times h},\; l=1,\ldots,L-1$ and $W^{(L)} \in \mathbb{R}^{h \times c}$. 
Similarly to GCNII, parameters $\alpha_l$ and $\beta_l$ need to be specified by the user or tuned as hyperparamters. As  with GCNII \citep{pmlr-v119-chen20v}, we also specify a predefined  $\alpha := \alpha_l \in (0,1)$ for all layers and  decaying $\beta_l=\log(\lambda/l +1) \approx \lambda/l$ where $\lambda$ is a predefined parameter.

The main advantage with AdaGPR compared to conventional graph convolution methods and GCNII is that it can learn how to apply convolution at each layer. It is obvious that when  $\mu^{(l)}_0 =1.0$ or $\mu^{(l)}_1 =1.0$ for all $l=1,\ldots,L$ AdaGPR is  equivalent to a multilayer residual network or  GCNII, respectively. Again, notice that AdaGPR has a generalized Pagerank  at each layer with aggregations and nonlinear activations compared to APPNP and  GPR-GNN. To our knowledge AdaGPR is the first graph convolution model to apply layer-wise adaptive   Pagerank in a deep graph convolution model.

We point out that AdaGPR has more learning parameters and hyperparamters than GCNII. In practice, we have found that we need to consider $K$  as a hyperparameter that needs to be selected during the training phase. The increased number of hyperparameters is a limitation of the proposed method.  We can also use a different $K$ for each layer, however, that may be impractical due to the large combinations of GPRs we may have to consider. Depending on the learning problem, we may also have to apply a separate weight decay for $v^{(l)}$. 
  

\subsection{Learning Sparse Solutions for GPR}
There are several ways to learn $\mu^{(l)}$ of \eqref{eq:AdaGPR} such that $\sum_{k=0}^{K-1} \mu_{k}^{(l)} =1$. One of the simplest methods is to use the Softmax, however, the resulting $\mu^{(l)}$ may not be sparse which would not give us the desired interpretable results. Variants of Softmax \citep{Martins2016} such as sphericalmax and sum-normalization  may lead to the same limitation of spraseness in addition to the difficulty of implementing the restriction $\sum_{k=0}^{K-1}\mu^{(l)}_k \neq 0$.  Another approach is  message passing as  used in  GPR-GNN \citep{chien2021adaptive}, however, it can be computationally expensive to implement message passing in a deep GCN model such as our proposed method. 

We adopt the recently developed sparse activation function Sparsemax \citep{Martins2016}  for the task of predicting each $\mu^{(l)}$.
Without loss of generality we restate $\mu^{(l)}$ belonging to a $(K-1)$-dimensional simplex  $\Delta^{K-1} := \{\mu^{(l)} \in \mathbb{R}^{K} | \mathbf{1}^{\top}\mu^{(l)} = 1, \mu^{(l)} \leq \mathbf{0}\}$, then Sparsemax is the solution of
\begin{equation}
\mathrm{sparsemax}(z^{(l)}) =\underset{\mu^{(l)} \in \Delta^{K-1}}{\mathrm{argmin}} \|\mu^{(l)} - z^{(l)}  \|^{2}. \label{eq:sparsemax}
\end{equation}
The closed-form solution of \eqref{eq:sparsemax} is given by  $\mathrm{sparsemax}_{i}(z) = [z_i - \tau(z)]_{+}$ \citep{Martins2016}, where $\tau(z) = \frac{(\sum_{j\in k(z)} z_{(j)} ) - 1}{k(z)}$ with $k(z):= \max \{k \in [K]| 1 + k z_{(k)} > \sum_{j < k} z_{(j)} \}$ given  sorted $z_{(1)} \geq z_{(2)} \geq \ldots \geq z_{(K)}$. By empirical evaluations, we found that we can obtain better solutions for AdaGPR  by using $\exp(\mu^{(l)})$ instead of $\mu^{(l)}$, which  resembles a sparse version of softmax. 
Our implementations of AdaGPR use the Pytorch code for sparsemax associate with the paper \citep{Martins2016}\footnote{https://github.com/KrisKorrel/sparsemax-pytorch}. 

\section{Theoretical Analysis}\label{sec:TheoreticalAnalysis}
We give a new generalization error bound for the proposed method. 
Unlike existing bounds, our bound fully incorporates the information of the spectrum of the normalized adjacency matrix $\tilde{A}$ and thus can take the effect of oversmoothing into account.

We analyse generalization bounds under transductive settings \citep{10.5555/1641503.1641508,OonoS20}  for semi-supervised node classification. We recall that $X \in \mathbb{R}^{N \times q}$ is the feature matrix of $N$ nodes with an associated  graph $G = (V,E)$  and consider a 1-class labeled output $Y \in  \mathbb{R}^{N \times 1}$. Let us consider the  sets $\mathcal{X}$ and $\mathcal{Y}$ such that $X \subseteq\mathcal{X}$, $Y \subseteq\mathcal{Y}$ and $(x_i ,y_i) \in \mathcal{X} \times \mathcal{Y}$. Let us consider $D_{\mathrm{train}}$ and $D_{\mathrm{test}}$ as the training and test sets, respectively. Samples are drawn without replacement from $D_{\mathrm{train}}$ and $D_{\mathrm{test}}$ such that $D_{\mathrm{train}} \cup D_{\mathrm{test}} = V$ and  $D_{\mathrm{train}} \cap D_{\mathrm{test}} = \emptyset$. Given $M: =|D_{\mathrm{train}}|$ and $U: =|D_{\mathrm{test}}|$, we define $Q:=1/M + 1/U$. Let $\mathcal{F} \subset \{\mathcal{X} \rightarrow \mathcal{Y} \}$ be the hypothesis for the transductive learning for AdaGPR. For a predictor $h : \mathcal{X} \rightarrow \mathcal{Y},\;\; h \in \mathcal{F}$ and  a loss function $l(\cdot,\cdot)$ (e.g., sigmoid, sigmoid cross entropy),  we denote the training error by $R(h) = \frac{1}{M}\sum_{n \in V_{\mathrm{train}}}l(h(x_{n}),y_{n})$ and test error by $\hat{R}(h) = \frac{1}{U}\sum_{n \in V_{\mathrm{test}}}l(h(x_{n}),y_{n})$. Using a well-known result from \citet{10.5555/1641503.1641508}, for a given hypothesis class $\mathcal{F}$ we state the generalization bounds based on transductive Rademacher complexity  $\mathcal{R}(\mathcal{F},p)$ with $p \in [0,0.5]$ and $S := \frac{2(M+U)\min(M,U)}{(2(M+U)-1)(2\min(M,U) -1)}$ and probability $1-\delta$ as\footnote{Here, by abuse of notation, we regard $\mathcal{F}$ 
as a subset of $\mathbb{R}^N$ by the identity $\{f(X) \mid f\in \mathcal{F}\} \subset \mathbb{R}^N$ although it is a set of functions from $\mathcal{X}$ to $\mathcal{Y}$.}  
\begin{equation}
R(h) \leq \hat{R}(h) + \mathcal{R}(\mathcal{F},p_0) + c_0Q \sqrt{\min(M,U)} +  \sqrt{\frac{SQ}{2}\log{\frac{1}{\delta}}},\label{eq:trans_gen_bound}
\end{equation} 
where
\begin{equation*}
\mathcal{R}(\mathcal{V},p) = Q \mathbb{E}_{\epsilon}\bigg[ \sup_{v \in \mathcal{V}} \langle \epsilon , v \rangle \bigg],
\end{equation*}
where $\epsilon = (\epsilon_1,\ldots,\epsilon_N)$ is a sequence of i.i.d. Rademacher variables with distribution $\mathbb{P}(\epsilon_i=1) = \mathbb{P}(\epsilon_i = -1) =p$ and $\mathbb{P}(\epsilon_i = 0) = 1-2p$ and $c_0$ is a constant. Following \citet{OonoS20}, the generalization error bound holds for the special case of $p=p_0 = MU/(M+U)^2$.  

For the ease of analysis, we consider unscaled weight in \eqref{eq:AdaGPR} with $\beta_l = 1.0$ and a single $\alpha_l = \alpha \in (0,1)$ for all layers.
We consider a predefined $\mu^{(l)} \in [0,1]^{K}$ with $\sum_{k=0}^{K-1} \mu_k^{(l)}=1$ for each layer $l$ to construct layer-wise a GPR  as $\tilde{A}(\mu^{(l)}) := \sum_{k=0}^{K-1} \mu^{(l)}_k \tilde{A}^{k}$.
Let us define $C_0,\ldots,C_{L} \in \mathbb{N}_{+}$ with $C_0 = q$, $C_1 = \cdots =C_{L-1} = h$ and $C_{L} = 1$ to represent the dimensions of hidden layers and the output of AdaGPR.
We define the hypothesis class for  AdaGPR fr semi-supervised node-classification as
\begin{align}
 \mathcal{F} = \Big\{ X \mapsto 
f^{(L)} \circ \dots \circ f^{(1)}(X)
~\Big|~  f^{(l)}(\cdot) &= \sigma\left(((1-\alpha)\tilde{A}(\mu^{(l)})(\cdot) + \alpha \sigma(X W^{(0)}))W^{(l)}\right),
\notag\\
& \|W_{\cdot c}^{(l)} \|_{1} \leq  B^{(l)}\; 
 \mathrm{for\; all} \; c \in [C_{l+1}] \Big\}, \label{eq:hypo_AdaGPR}
\end{align}
where $W^{(l)} \in \mathbb{R}^{C_l \times C_{l+1}} \; l = 0,\ldots,L$, and $\sigma : \mathbb{R} \rightarrow \mathbb{R}$ is a $1$-Lipschitz function such that $\sigma(0) = 0$ with bounded output\footnote{This is just a technical condition to ensure the input to each layer is bounded.} as $|\sigma(\cdot)| \leq R$, and $B^{(l)}\; l = 0,\ldots,L$ are constants. We point out that $\sigma(\cdot)$ can be a ReLU ( with output clipping) or a sigmoid function.

Analysing the Rademacher complexity allows up to obtain a data dependent bounds for our proposed model. Theorem 1 gives the Rademacher complexity for the AdaGPR.

\begin{thm}\label{thm:GeneralizationBound}
Given the hypothesis class $\mathcal{F}$,  the Rademacher complexity of the AdaGPR is bounded by
\begin{align} 
 \mathcal{R}(\mathcal{F},p_0)   \leq & Q C^{'} \Bigg\{\sqrt{\frac{2MU}{(M+U)^2}}B^{(0)}\alpha\Bigg[\sum_{l=1}^{L} (1-\alpha)^{l} 2^{l} \prod_{j=0}^{l-1} B^{(L-j)}\bigg(\sum_{i=1}^{N} \sum_{k=0}^{K-1} \mu_{k}^{(L-j)}  |\lambda_{i}|^{k} \bigg)  \| X\|_{\mathrm{F}} \Bigg]\notag \\
&~~~+ 
 \sum_{l=1}^{L} (1-\alpha)^{l+1} 2^{l} \times  \prod_{j=0}^{l} \bigg[ B^{(L-j)} \bigg(\sum_{i=1}^{N} \sum_{k=0}^{K-1} \mu_{k}^{(L-j)}  |\lambda_{i}|^{k} \bigg)\bigg] D\Bigg\},  
\label{eq:main_gen_gcnII_gpr} 
\end{align}
where $\lambda_{i}$ is the $i$th largest eigenvalue of $\tilde{A}$, $D=\sqrt{N}R$, and $C^{'}$ is a universal constant.
\end{thm}

We  extend the hypothesis class \eqref{eq:hypo_AdaGPR} to derive the  hypothesis class for GCNII by setting $\mu_{1}^{(l)} = 1.0, \; l=1,\ldots,L$ and obtain the Rademacher complexity for GCNII given in the Corollary 1. 

\begin{col}
The Rademacher complexity of the GCNII is bounded as 
\begin{multline} 
\mathcal{R}(\mathcal{F},p_0)   \leq 
Q C^{'}\Bigg\{\sqrt{\frac{2MU}{(M+U)^2}} B^{(0)} \alpha \sum_{l=1}^{L} 2^{l} (1-\alpha)^{l}
\prod_{j=0}^{l-1} B^{(L-j)}\bigg(\sum_{i=1}^{N} |\lambda_{i}|  \bigg)  \| X \|_{\mathrm{F}}   
+ ~~~~~~~~~~~~~~~~~~~~~~~~~~\\
\sum_{l=1}^{L} (1-\alpha)^{l+1} 2^{l} 
\prod_{j=0}^{l} \Bigg[ B^{(L-j)}\bigg(\sum_{i=1}^{N} |\lambda_{i}| \bigg)  \Bigg]D\Bigg\}, \label{eq:main_gen_gcnII} 
\end{multline}
where $\lambda_{i}$ is the $i$th largest eigenvalue of $\tilde{A}$, $D=\sqrt{N}R$, and $C^{'}$ is a universal constant.
\end{col}

The proof is given in Appendix \ref{sec:ProofGenBound}.
We notice that the bounds \eqref{eq:main_gen_gcnII_gpr} and \eqref{eq:main_gen_gcnII} are characterized by the spectrum of $\tilde{A}$. It shows that the mixing speed of information by node aggregations at each layer affects the model complexity. 
The use of the normalized adjacency matrix results in a eigenvalue spectrum of $1 = \lambda_1 \geq \lambda_2 \geq \cdots\geq \lambda_N \geq -1 $ and as $k$ increases the summations of the eigenvalue spectrum with the higher powers shrink quickly.
With a large $k$, the Rademacher complexity may become small which coincides with the intuition that the oversmoothing effect makes the model ``simpler'' and gives smaller generalization gap. 
Our bound successfully characterizes such an effect through the spectrum information which represents how fast the node features are mixed by aggregation. 
On the other hand, multiple applications of node aggregations would induce strong oversmoothing and results in underfitting (while the generalization gap is small). 
Because of it, the model complexity is uniformly bounded even there are many additional multiple node aggregation terms. 
Our proposed method automatically finds the appropriate weight that fits the data well. 
It is also important to notice that the deeper layers have a strong influence on the overall generalization bound due to recursive summations, which is coming from the input injection ($H^{(0)}$) to every layer. This indicates that the recursive multiplications of the spectral components in deeper layers induce stronger bias (although it simultaneously yields smaller generalization gap). Hence, in order to have less oversmoothing and to have a small overall bias less graph convolutions are preferred at deep layers. This observation agrees with the experimental results (Table \ref{table:cornell}).

The characterization by the spectrum is beneficial especially for large node-degree graphs. 
Indeed, a PAC-Bayesian bound for GCNs given by \citet{DBLP:journals/corr/abs-2012-07690} includes $d^{k/2}$ instead of $|\lambda_i|^k$ where $d$ is the maximum node degree of the graph. 
Such a bound becomes loose for large degree $d$. However, a graph with large degree likely to have small spectrum $\lambda_i$ (because it can ``mix'' the information rapidly) and thus our bound gives a tighter bound, which is contrary to the existing bound. 
Other bounds (e.g., by \citet{OonoS20}) are merely characterized by the spectral norm of the weight matrices, but our bound is characterized by not only the spectral norm $B$ but also the spectrum of the node aggregation.

\begin{table*}[t]
\centering
\begin{tabular}{l|c|c|c|c|c|c|c}\hline
Properties & Cora  & Citeseer & Pubmed & Chameleon & Cornell & Texas  & Wisconsin  \\ \hline
Classes & 7 & 4 &  3 &4 & 5 &5  &5  \\
Nodes & 2708 & 3327  & 19717 & 2277 & 183 & 183 & 251 \\
Edges & 5429 & 4732  & 44338 &36101 & 295 & 309 & 499 \\
Features & 1433 & 3703  & 500 &2325 & 1703 & 1703 & 1703 \\
\hline
\end{tabular}
\caption{Properties of datasets used for node-classification} 
 \label{table:datasets}
\end{table*}

\section{Experiments}
In this section we discuss node classification experiments that we carried out to evaluate AdaGPR. Additionally, we discuss the behaviour of layer-wise sparse solutions of generalised Pagerank coefficients to understand the adaptive behaviour of AdaGPR.

\begin{table*}[t]
\vspace{-0.5cm}
\centering
\begin{tabular}{l|l|llllll}\hline
\multicolumn{1}{l|}{\multirow{2}{*}{Dataset}} & \multirow{2}{*}{Method} & \multicolumn{6}{l}{ Layers} \\
\multicolumn{1}{l|}{}                  &                   & {\small  2 }& {\small 4  }& {\small 8  }& {\small 16  }& {\small 32  }& {\small 64} \\ \hline
\multirow{6}{*}{Cora}                  & {\small  GCN              }& {\small \bf{81.1 } }& {\small 80.4  }& {\small 69.5   }& {\small 64.9  }& {\small 60.3  }& {\small 28.7 }\\
                                       & {\small  GCN(Drop)        }& {\small \bf{82.8}  }& {\small 82.0  }& {\small 75.8  }& {\small 75.7  }& {\small 62.5  }& {\small 49.5 }\\
                                       & {\small  JKNet            }& {\small -  }& {\small 80.2  }& {\small \bf{80.7}  }& {\small 80.2  }& {\small 81.1  }& {\small 71.5 }\\
                                       & {\small  JKNet(Drop)      }& {\small -  }& {\small \bf{83.3}  }& {\small 82.6  }& {\small 83.0  }& {\small 82.5  }& {\small 83.2} \\
                                       & {\small  Incep            }& {\small -  }& {\small 77.6  }& {\small 76.5  }& {\small  81.7 }& {\small  \bf{81.7} }& {\small 80.0 }\\
                                       & {\small  Incep(Drop)      }& {\small -  }& {\small 82.9  }& {\small 82.5  }& {\small 83.1  }& {\small   83.1 }& {\small \bf{83.5}  } \\
                                       & {\small  GCNII  (hidden 64)          }& {\small 82.2  }& {\small 82.6  }& {\small 84.2  }& {\small 84.6  }& {\small   85.4 }& {\small \bf{85.5} }\\
                                       & {\small  GCNII* (hidden 64)       }& {\small  80.2 }& {\small 82.3  }& {\small 82.8  }& {\small 83.5  }& {\small 84.9   }& {\small \bf{85.3}   }\\
                                       & {\small  AdaGPR (hidden 32, GPR coeffs. 4)       }& {\small 83.8 }& {\small 84.5  }& {\small 84.8  }& {\small 85.0  }& {\small \bf{85.0}  }& {\small 85.0 }\\ \hline
\multirow{3}{*}{Citeseer}              & {\small  GCN              }& {\small \bf{70.8}  }& {\small 67.7  }& {\small 30.2  }& {\small 18.3  }& {\small 25.0  }& {\small 20.0 }\\
                                       & {\small  GCN(Drop)        }& {\small \bf{72.3}  }& {\small 70.6  }& {\small 61.4  }& {\small 57.2  }& {\small 41.6  }& {\small 34.4 }\\
                                       & {\small  JKNet            }& {\small - }& {\small 68.7  }& {\small 67.7  }& {\small \bf{69.8}  }& {\small 68.2  }& {\small 63.4 }\\
                                       & {\small  JKNet(Drop)      }& {\small - }& {\small 72.6  }& {\small 71.8  }& {\small 72.6  }& {\small 70.8  }& {\small 72.2}\\
                                       & {\small  Incep            }& {\small - }& {\small 69.3  }& {\small 68.4  }& {\small 70.2  }& {\small \bf{72.6}  }& {\small 71.0}\\
                                       & {\small  Incep(Drop)      }& {\small - }& {\small \bf{72.7}  }& {\small 71.4  }& {\small 72.5  }& {\small 72.6  }& {\small 71.0 }\\
                                       & {\small  GCNII (hidden 256)           }& {\small 68.2  }& {\small 68.9  }& {\small 70.6  }& {\small 72.9  }& {\small \bf{73.4}  }& {\small 73.4 }\\
                                       & {\small  GCNII* (hidden 256)       }& {\small 66.1  }& {\small  67.9 }& {\small  70.6 }& {\small 72.0  }& {\small \bf{73.2}  }& {\small 73.1 }\\
                                       & {\small  AdaGPR (hidden 64, GPR coeffs. 16)       }& {\small 59.9  }& {\small 68.6  }& {\small 73.2  }& {\small \bf{73.5}  }& {\small 73.4  }& {\small 73.1} \\ \hline
\multirow{3}{*}{Pubmed}              & {\small  GCN              }& {\small \bf{79.0}  }& {\small 76.5  }& {\small 60.1  }& {\small 40.9  }& {\small 22.4  }& {\small 35.5 }\\
                                       & {\small  GCN(Drop)        }& {\small \bf{79.6}  }& {\small 79.4  }& {\small 78.1  }& {\small 78.5  }& {\small 77.0  }& {\small 61.5 }\\
                                       & {\small  JKNet            }& {\small - }& {\small 78.0  }& {\small \bf{78.1}  }& {\small 72.6  }& {\small 72.4  }& {\small 74.5} \\
                                       & {\small  JKNet(Drop)      }& {\small - }& {\small 78.7  }& {\small 78.7  }& {\small 79.1  }& {\small \bf{79.2}  }& {\small 78.5}\\
                                       
                                       & {\small  IncepGCN            }& {\small - }& {\small 77.7  }& {\small 77.9  }& {\small 74.9  }& {\small OOM  }& {\small OOM}\\
                                       & {\small  IncepGCN(Drop)      }& {\small - }& {\small \bf{79.5}  }& {\small 78.6  }& {\small 79.0  }& {\small OOM  }& {\small OOM }\\
                                       & {\small  GCNII (hidden 256)           }& {\small 78.2  }& {\small 78.8  }& {\small 79.3  }& {\small \bf{80.2}  }& {\small 79.8  }& {\small 79.7 }\\
                                       & {\small  GCNII* (hidden 256)       }& {\small 77.7  }& {\small  78.2 }& {\small  78.8 }& {\small \bf{80.3}  }& {\small 79.8  }& {\small 80.1} \\
                                       & {\small  AdaGPR (hidden  128, GPR coeffs. 4)       }& {\small  78.3  }& {\small 78.8  }& {\small  79.4 }& {\small \bf{79.6} }& {\small 79.3 }& {\small OOM} \\                                  
\hline\end{tabular}
\caption{Accuracy for semi-supervised node classification} \label{table:semi}
\vspace{-0.2cm}
\end{table*}

\subsection{Setup}
We performed semi-supervised and fully-supervised node classification. Datasets and their properties used in our experiments are listed in Table \ref{table:datasets}. Since our  method stems from  GCNII, we used a similar  experimental setting as  in \citet{pmlr-v119-chen20v} and borrowed their reported results for baseline methods. In addition to the hyperparameters $\alpha_l$, $\lambda_l$, weight decays, and dropout rates common with GCNII, the number of GPR coefficients $K$  and in some cases (semi-supervised learning) weight decay for learning weights $v^{(l)}$   in \eqref{eq:AdaGPR} are considered as hyperparameters.  We tuned hyperparameters based on the loss over the validation sets. The optimization method for all experiments is Adam with learning rate of $0.01$. We use the publicly available processed data provided by \citet{pmlr-v119-chen20v}. Further, we use code from \citet{pmlr-v119-chen20v} to assist our implementations.  The data and Pytorch based implementation of AdaGPR is available at \href{https://github.com/kishanwn/AdaGPR}{https://github.com/tophatjap/adaGPR}.
We carried out experiments on NVidia V100-PCIE-16GB GPUs hosted on Intel Xeon Gold 6136 processor servers. 

\subsection{Semi-Supervised Node Classification}
We used the commonly used citation datasets Cora, Citeseer, and Pubmed to evaluate performance of semi-supervised node classification. 
These datasets are split based on  the commonly used the setting in \citet{yang16-icml} that results in training sets with $20$ nodes per each class, test sets with $500$ nodes, and  validation sets with $1000$ nodes. The number of  coefficients of the GPR  is considered a  hyperparameter and selected  from $(2,3,4,8,16)$. We used the same hyperparameter ranges as in \citet{pmlr-v119-chen20v} for $\lambda$,  and dropout rates from $(0.1,\ldots,0.9)$. We fixed $\alpha=0.1$ following \citet{pmlr-v119-chen20v}.

We used separate weight decay rates for different learning weights in AdaGPR; $WD_1 \in (1.0,0.1,0.01,\ldots,0.0001)$ for $W^{(0)}$, $WD_2 = 0.0001$ for $W^{(l)},\; l=1,\ldots,L$, and $WD_3 \in (1.0,0.1,0.01)$ for $v^{(l)}\; l=1,\ldots,L$. $WD_1$ and $WD_3$ are selected from hyperparameter tuning (see Section 1 of the supplementary materials section for details).
We borrowed results for baseline methods  Vanilla GCN \citep{Kipf:2016tc}, JKNet \citep{pmlr-v80-xu18c}, IncepGCN \citep{DBLP:conf/iclr/RongHXH20}, and  GCNII \citep{pmlr-v119-chen20v} from \citet{pmlr-v119-chen20v}.

The Table \ref{table:semi} shows that classification accuracies for Cora using  AdaGPR did not out-perform the accuracy produced by GCNII. However, AdaGPR has produced better performances for shallow networks with layers ranging from 2 to 16 compared to GCNII.
AdaGPR achieved a slightly improved accuracy for Citeseer compared to GCNII. The noteworthy observation is that AdaGPR provides the best accuracy of $73.5$ with 16  layers  and 32 hidden units compared to the GCNII which used 32 layers and 256 hidden units. AdaGPR obtained a slightly lower accuracy for Pubmed compared to GCNII. 
The stable accuracies with the increase in depth for all  datasets  show robustness against oversmoothing of AdaGPR.

\begin{wraptable}{l}{0.46\textwidth}
\vspace{-0.3cm}
\sbox0{
\begin{tabular}{l|ccccccc}\hline
\multirow{2}{*}{Method} & \multicolumn{7}{l}{Dataset} \\ \cline{2-8} 
                  & Cora & Citeseer & Pubmed &  Chameleon  & Cornell & Texas & Wisconsin  \\  \hline
{\small GCN }               & {\small 85.77} & {\small 73.68} & {\small 88.13} &  {\small 28.18} & {\small 52.70} & {\small 52.16} & {\small 45.88}  \\
{\small GAT  }              & {\small 86.37 }& {\small 74.32} & {\small 87.62} &  {\small 42.93 }&  {\small 54.32 }&{\small 58.38 } &  {\small 49.41 }\\
{\small Geom-GCN-I     }      & {\small 85.19 }& {\small \bf{77.99} }&  {\small 80.05} & {\small 60.31 }& {\small 56.76 }& {\small 57.58 }& {\small 58.24 } \\
{\small APPNP   }             &{\small  87.87 }&{\small  76.53} & {\small 89.40} &  {\small   54.3 }&{\small  73.51 }&{\small  65.41 }&  {\small 69.02} \\
{\small JKNet   }             &{\small  85.25 (16) }&{\small  75.85 (8)} & {\small 88.94 (64)} &  {\small  60.07 (64) }&{\small  57.30 (4) }& {\small 56.49 (32) }&{\small  48.82 (8) }\\
{\small JKNet(Drop) }         &{\small  87.46 (16) }&{\small  75.96 (8) }& {\small 89.45 (64)} &  {\small   62.08 (64) }&{\small  61.08 (4) }& {\small 57.30 (32)}&{\small   50.59 (8) }\\
{\small IncepGCN(Drop)   }      &{\small  86.86 (8) }& {\small 76.83 (8) }& {\small 89.18} &  {\small  61.71 (4) }&{\small  61.62 (16) }& {\small 57.84 (8)}&{\small  50.20 (8)  } \\
{\small GPR-GNN  }       & {\small 88.16 }& {\small 77.39 } & {\small 85.8} &  {\small 63.22} & {\small 78.37} & {\small 77.30} & {\small 81.57 }  \\
{\small GCNII  }        & {\small \bf{88.49} (64)} & {\small 77.08 (64)} & {\small 89.57 (64)} &  {\small 60.61 (8) }& {\small 74.86 (16)} & {\small 69.46 (32) }&  {\small 74.12 (16) }\\
{\small GCNII* }          & {\small 88.01 (64) }& {\small 77.13 (64) }& {\small \bf{90.30 (64)}} &  {\small 62.48 (8) }& {\small 76.49 (16) }& {\small 77.84 (32)} &  {\small 81.57 (16)} \\
{\small AdaGPR  }         & {\small 88.19 (64,3)} & {\small 77.25 (64,4)} & {\small \bf{90.23 (4,3)}} &   {\small \bf{64.71} (2,3)} & {\small \bf{82.70} (4,2)}  & {\small \bf{81.08} (4,4)}   &  {\small \bf{83.53} (16,3)}  \\
\hline
\end{tabular}
}
\rotatebox{90}{%
\parbox{\wd0}{\centering\usebox0
\caption{Accuracy for fully-supervised node classification} \label{table:full}
\endgraf\bigskip
}}
\vspace{-1cm}
\end{wraptable}

\subsection{Fully-Supervised Node Classification}

We experimented with fully-supervised node classification using the standards baseline graph datasets of Cora, Citeseer, Pubmed, Chameleon, Cornell, Texas, and Wisconsin. As suggested in \citet{Pei-20-Geom-GCN},  all these datasets were randomly split into  training, validation and testing sets consisting of nodes by each class with percentages of $60\% $, $20\% $, and $20\% $, respectively. We ran experiments over 10 different random splits as used in \citet{pmlr-v119-chen20v}.  For fair comparisons with \citet{pmlr-v119-chen20v}  we used the 64 hidden units for all methods. Hyperparameter sets for dropout rates, and $K$  are same as fully-supervised learning. Similar to \citet{pmlr-v119-chen20v}, we used a single weight decay selected from the set $(0.001,0.0005,\ldots,1\mathrm{e-}6)$, $\alpha \in (0.1,\cdots,0.9)$, and $\lambda \in (0.5,1.0,1.5)$.

 The mean accuracy for node classification of AdaGPR and baseline methods (borrowed from  \citet{pmlr-v119-chen20v}) are shown in the Table \ref{table:full}.  These baseline methods are Vanilla GCN \citep{Kipf:2016tc}, GAT \citep{velikovi2017graph}, Geom-GCN \citep{Pei-20-Geom-GCN}, APPNP \citep{KlicperaBG19}, JKNet \citep{pmlr-v80-xu18c}, IncepGCN \citep{DBLP:conf/iclr/RongHXH20}, and GCNII \citep{pmlr-v119-chen20v}.  We also experimented with GPR-GNN whose results are included in Table  \ref{table:full}. In addition to accuracy of AdaGPR, we show the number of layers and  number of GPR coefficients ($K$) in brackets that were selected from the hyperparameter tuning.

Form Table \ref{table:full} we can see that AdaGPR has obtained comparable accuracies  compared to GCNII for  Cora, Citeseer, and Pubmed.   
Chameleon dataset has a similar number of nodes as with Cora and Citeseer (Table \ref{table:datasets}), however, it  has a larger  number of edges compared to Cora and Citeseer. This indicates that Chameleon has a dense adjacency matrix compared to Cora and Citeseer, which may lead to faster oversmoothing with multiple convolutions.
 This observation is reflected in AdaGPR model with $2$ layers and $3$ coefficients giving the best accuracy for Chameleon. Notice that GCNII also has used a smaller network (8 layers) for Chemeleon compared to other datasets. Further, it is noteworthy that GPR-GNN which is another shallow model has gained a accuracy comparable to AdaGPR for Chemeleon.

There is a significant high accuracy  for the three small scale datasets of Cornell, Texas, and Wisconsin with AdaGPR compared to all the baseline methods. Again, we can see that the increased performance with AdaGPR are achieved for Cornell and Texas with less number of convolution layers compared to GCNII. These observations   provide evidence that adaptive GPR  can perform model compression while enhancing prediction accuracy.

\subsection{Layer-wise GPR Adaptation}
We can quantitatively understand the amount of convolution by different orders of the normalized adjacency matrix at each layer by analysing the coefficients of each generalized Pagerank. In order to demsntrate layer-wise adaptation, we show coefficients of each generalized Pagerank at each layer for Cornell in Table \ref{table:cornell}. Notice the clear lawer-wise adaptation where only the first two layers apply  graph convolutions with gradual decrease of the GPR from shallow layers to deeper layers and the last two layers of the trained model have no graph convolution.

\begin{table}
\centering
\begin{tabular}{l|cc}\hline
\multirow{2}{*}{Layers} & \multicolumn{2}{l}{GPR Coeff.} \\ \cline{2-3} 
  & {\small 0  }& {\small 1  } \\ \hline
{\small 1} & {\small 0.5150} & {\small 0.4849} \\
{\small 2} & {\small 0.8581} & {\small 0.1418} \\
{\small 3 }& {\small 1} & {\small 0} \\
{\small 4} & {\small 1 }& {\small 0} \\
\hline\end{tabular}
\caption{GPR coefficients of Cornell}\label{table:cornell}
\end{table}

\section{Conclusions}
We proposed the AdaGPR to  perform layer-wise adaptive graph convolution using generalized Pageranks within GCNII models. 
 We provide generalization bounds to analyse the relationship between eigenvalue spectrum of a graph and the depth of the network and its effect on oversmoothing. We evaluate our proposed method using benchmark node-classification datasets to show performance improvements compared to other GCN models.  By analysing coefficients of the generalized Pagerank in the trained models, we confirm that adaptive behaviour of graph convolution in each layer.

\bibliographystyle{apalike}
\bibliography{ref}

\appendix

\section{Proofs of Generalization Bounds}\label{sec:ProofGenBound}
In this section we provide detailed proofs of Theorems given in the Section \ref{sec:TheoreticalAnalysis}.
The following transductive Rademacher complexity is defined in \citet{10.5555/1641503.1641508}. 

\begin{definition}
Given $p \in [0,0.5]$ and $\mathcal{V} \subset \mathbb{R}^{N}$, the transductive Rademacher complexity is defined as
\begin{equation*}
\mathcal{R}(\mathcal{V},p) = Q \mathbb{E}_{\epsilon}\bigg[ \sup_{v \in \mathcal{V}} \langle \epsilon , v \rangle \bigg],
\end{equation*}
where $Q = \frac{1}{M} + \frac{1}{N}$ and $\epsilon = (\epsilon_1,\ldots,\epsilon_N)$ is a sequence of i.i.d. Rademacher variables with distribution $\mathbb{P}(\epsilon_i=1) = \mathbb{P}(\epsilon_i = -1) =p$ and $\mathbb{P}(\epsilon_i = 0) = 1-2p$. 
\end{definition}

Below we restate the symmetric Rademacher complexity \citep{OonoS20}, a variant of the above tranductive Rademacher complexity.

\begin{definition}
Given $p \in [0,0.5]$ and $\mathcal{V} \subset \mathbb{R}^{N}$, the symmetric transductive Rademacher complexity is defined as
\begin{equation*}
\mathcal{\bar{R}}(\mathcal{V},p) = Q \mathbb{E}_{\epsilon}\bigg[ \sup_{v \in \mathcal{V}} |\langle \epsilon , v \rangle | \bigg],
\end{equation*}
where $Q = \frac{1}{M} + \frac{1}{N}$ and $\epsilon = (\epsilon_1,\ldots,\epsilon_N)$ is a sequence of i.i.d. Rademacher variables with distribution $\mathbb{P}(\epsilon_i=1) = \mathbb{P}(\epsilon_i = -1) =p$ and $\mathbb{P}(\epsilon_i = 0) = 1-2p$. 
\end{definition}  

In \citep{OonoS20}, it has been shown that $\mathcal{R}(\mathcal{V},p) \leq \mathcal{\bar{R}}(\mathcal{V},p)$.


Below we provide the proof for the Theorem 1.

\textit{Proof of Theorem 1.}{
We use the symmetric Rademacher complexity 
\begin{equation}
\mathcal{\bar{R}}(\mathcal{F},p) =  \mathbb{E}_{\boldsymbol{\epsilon}}  \bigg[\sup_{v \in \mathcal{F}} |\langle \boldsymbol{\epsilon}, v \rangle | \bigg],
\end{equation}
which upper bounds the Rademacher complexity $\mathcal{R}(\mathcal{F},p)$ in (9) as  $\mathcal{R}(\mathcal{F},p) \leq \mathcal{\bar{R}}(\mathcal{F},p)$. 
We give the bound for a general $p$. The assertion can be obtained by substituting $p \leftarrow p_0$.

In the rest of the proof we abbreviate  row $s$ of any matrix $Z$ by $Z_s := Z[s,:]$, columns $c$ by $Z_{\cdot c} := X[:,c]$, and an element by $Z_{sc} := Z[s,c]$. For the convenience of analysis we break the hypothesis class $\mathcal{F}$ in (10) 
into different components and define  
\begin{gather*}
\mathcal{H}^{(0)} = \Bigg\{ \sum_{c=1}^{C_0} X_{\cdot c}w^{(0)}_{c} |  \| w^{(0)}_{c} \|_{1} \leq B^{(0)}  \Bigg\},\\
\mathcal{\tilde{H}}^{(0)} = \sigma \circ \mathcal{H}^{(0)}, \\
\mathcal{H}^{(l+1)} = \Bigg\{ \sum_{c=1}^{C_{l+1}} ((1-\alpha)[\tilde{A}(\mu^{(l)})Z]_{\cdot c} +  \alpha H_{\cdot c})w_{c}^{(l)} | Z_{\cdot c} \in \mathcal{\tilde{H}}^{(l)}, H_{\cdot c} \in \mathcal{\tilde{H}}^{(0)}, \| w^{(l)} \|_{1} \leq B^{(l)}  \Bigg\}, \\
\mathcal{\tilde{H}}^{(l +1)} = \sigma \circ \mathcal{H}^{(l)} \; l = 1,\ldots,L.
\end{gather*}

Now, for a given layer $l+1$, we have 
\begin{equation} 
\begin{split}
Q^{-1}\mathcal{\bar{R}}(\mathcal{\tilde{H}}^{(l+1)}, p)  &= \mathbb{E}_{\boldsymbol{\epsilon}}  \bigg[\sup_{\|w^{(l)} \|_1 \leq B^{(l)}, Z_{\cdot c} \in \mathcal{H}^{(l)}, H_{\cdot c} \in \mathcal{\tilde{H}}^{(0)}  } \bigg| \sum_{n=1}^{N} \epsilon_{n} \sum_{c=1}^{C_{l+1}} ((1-\alpha)[\tilde{A}(\mu^{(l)})Z]_{nc} + \alpha  H_{nc})w_{c}^{(l)}) \bigg| \bigg]  \\
 & = \mathbb{E}_{\boldsymbol{\epsilon}}  \bigg[ \sup_{ \|w^{(l)} \|_1 \leq B^{(l)}, Z_{\cdot c} \in \mathcal{H}^{(l)}, H_{\cdot c} \in \mathcal{\tilde{H}}^{(0)} } \bigg|  \sum_{c=1}^{C_{l+1}} \sum_{n=1}^{N} \epsilon_{n} ((1-\alpha)[A(\mu^{(l)})Z]_{nc} + \alpha  H_{nc})w_{c}^{(l)}) \bigg| \bigg] \\
 & =B^{(l)} \mathbb{E}_{\boldsymbol{\epsilon}}  \bigg[\sup_{Z \in \mathcal{H}^{(l)}, H \in \mathcal{\tilde{H}}^{(0)}} \bigg|   \sum_{n=1}^{N} \epsilon_{n} ((1-\alpha)[\tilde{A}(\mu^{(l)})Z]_{n} + \alpha  H_n) \bigg| \bigg] \\
  & =B^{(l)} \mathbb{E}_{\boldsymbol{\epsilon}}  \bigg[\sup_{Z \in \mathcal{H}^{(l)}, H \in \mathcal{H}^{(0)}} \bigg|   (1-\alpha) \sum_{n=1}^{N} \epsilon_{n} [\tilde{A}(\mu^{(l)})Z]_{n} +  \alpha \sum_{n=1}^{N} \sigma_{n}   H_n) \bigg| \bigg] \\
 &  = B^{(l)} (1 - \alpha)   \mathbb{E}_{\boldsymbol{\epsilon}}\bigg[\sup_{Z \in \mathcal{H}^{(l)}} \bigg|   \sum_{n=1}^{N} \epsilon_{n} [\tilde{A}(\mu^{(l)})Z]_{n} \bigg| \bigg]  + \alpha B^{(l)} \mathbb{E}_{\boldsymbol{\epsilon}}  \bigg[\sup_{H \in \mathcal{\tilde{H}}^{(0)} } \bigg| \sum_{n=1}^{N} \epsilon_{n}   H_n \bigg|\bigg].
\end{split} \label{eq:reduction_rad-step1}
\end{equation}

Let $\epsilon'=(\epsilon'_1,\dots,\epsilon'_N)$ be a random variable that 
is independent to and has the identical distribution as $\epsilon$.
Then, we have that 
\begin{align*}
& \mathbb{E}_\epsilon\left[\sup_{Z \in \mathcal{H}^{(l)}} \left|\sum_{n=1}^N\epsilon_n [\tilde{A}(\mu^{(l)}) Z]_n \right|\right]
= \mathbb{E}_\epsilon\left[\sup_{Z \in \mathcal{H}^{(l)}} \left|\sum_{n=1}^N\epsilon_n [\tilde{A}(\mu^{(l)}) \frac{1}{2p}\mathbb{E}_{\epsilon'}[\epsilon'\epsilon'^\top]Z]_n \right|\right]~~(\because \mathbb{E}_{\epsilon'}[\epsilon'\epsilon'^\top] = 2p I) \\
& \leq 
\frac{1}{2p}\mathbb{E}_{\epsilon,\epsilon'}\left[\sup_{Z \in \mathcal{H}^{(l)}} \left|\epsilon^\top \tilde{A}(\mu^{(l)}) \epsilon' \epsilon'^\top Z \right|\right] 
\leq 
\frac{1}{2p}\mathbb{E}_{\epsilon,\epsilon'}\left[\left|\epsilon^\top \tilde{A}(\mu^{(l)}) \epsilon'  \right| \sup_{Z \in \mathcal{H}^{(l)}} \left|\epsilon'^\top Z \right|\right] \\
& =
\frac{1}{2p}\mathbb{E}_{\epsilon'}\left[\mathbb{E}_{\epsilon}\left[\left|\epsilon^\top \tilde{A}(\mu^{(l)}) \epsilon'  \right| \right]\sup_{Z \in \mathcal{H}^{(l)}} \left|\epsilon'^\top Z \right|\right] \\
& \leq
\frac{1}{2p}\mathbb{E}_{\epsilon'}\left[\sqrt{\mathbb{E}_{\epsilon}\left[\left(\epsilon^\top \tilde{A}(\mu^{(l)}) \epsilon'  \right)^2 \right]} \sup_{Z \in \mathcal{H}^{(l)}} \left|\epsilon'^\top Z \right|\right] 
=
\frac{1}{2p}\mathbb{E}_{\epsilon'}\left[\sqrt{\mathbb{E}_{\epsilon}\left[\epsilon'^\top \tilde{A}(\mu^{(l)}) \epsilon \epsilon^\top \tilde{A}(\mu^{(l)}) \epsilon' \right]} \sup_{Z \in \mathcal{H}^{(l)}} \left|\epsilon'^\top Z \right|\right]
\\
& =
\mathbb{E}_{\epsilon'}\left[\sqrt{ \epsilon'^\top \tilde{A}(\mu^{(l)})^2 \epsilon'  } \sup_{Z \in \mathcal{H}^{(l)}} \left|\epsilon'^\top Z \right|\right],
\end{align*}
where we used $\mathbb{E}_{\epsilon}[\epsilon\epsilon^\top] = 2p I$ in the last equation. 
Here, by the Hanson-Wright concentration inequality (see, for example, Theorem 2.5 of \citep{adamczak2015note}) implies that 
$$
\mathbb{P}[|\epsilon'^\top \tilde{A}(\mu^{(l)})^2 \epsilon' - \mathbb{E}_{\epsilon'}[\epsilon'^\top \tilde{A}(\mu^{(l)})^2 \epsilon']| \geq c(\sqrt{2p} \|\tilde{A}(\mu^{(l)})^2\|_F \sqrt{ t} + \|\tilde{A}(\mu^{(l)})^2\| t)] \leq \exp(- t)~~~(t>0),
$$
with a universal constant $c$, where $\|A\|_F = \sqrt{\mathrm{Tr}[AA^\top]}$\footnote{There also exists a uniform type Hanson-Wright inequality.}.
Moreover, Talagrand's concentration inequality yields 
$$
\mathbb{P} \left[  \left| \sup_{Z \in \mathcal{H}^{(l)}} \epsilon'^\top Z \right|  \geq c'\left( \mathbb{E}_{\epsilon'}\left[\sup_{Z \in \mathcal{H}^{(l)}} \left| \epsilon'^\top Z \right| \right] + \sqrt{ N t \sup_{Z \in \mathcal{H}^{(l)}}\sum_{n=1}^N Z_n^2/N  } +   t  \sup_{Z \in \mathcal{H}^{(l)}}\|Z\|_\infty  \right)  \right]
\leq e^{-t}~~~(t > 0),
$$
where $c' > 0$ is a universal constant.
Then, by noticing that $\mathbb{E}_{\epsilon'}[\epsilon'^\top A \epsilon'] = 2p \mathrm{Tr}[A]$, these inequalities yield
\begin{align*}
& \mathbb{E}_{\epsilon'}\left[\sqrt{ \epsilon'^\top \tilde{A}(\mu^{(l)})^2 \epsilon'  } \sup_{Z \in \mathcal{H}^{(l)}} \left|\epsilon'^\top Z \right|\right] \\
\leq &
\int \sqrt{2p \mathrm{Tr}[\tilde{A}(\mu^{(l)})^2] + c (\sqrt{2p}\|\tilde{A}(\mu^{(l)})^2\|_F \sqrt{t} + \|\tilde{A}(\mu^{(l)})^2\|t) } \\
& \qquad\qquad\qquad\qquad\qquad\qquad\qquad\qquad c' ( | \mathbb{E}_{\epsilon'} \sup_{Z \in \mathcal{H}^{(l)}} \epsilon'^\top Z |  + \sqrt{t} \|\mathcal{H}^{(l)}\|_2 + t \|\mathcal{H}^{(l)}\|_\infty) 2 \exp(-t) \mathrm{d} t
\end{align*}
where we define $\|\mathcal{H}^{(l)}\|_{*} := \sup_{Z \in \mathcal{H}^{(l)}}\|Z\|_{*}$ for $*=2$ and $\infty$.
The right hand side can be further bounded as 
$
C \sqrt{\mathrm{Tr}[\tilde{A}(\mu^{(l)})^2]} \left( \mathbb{E}_{\epsilon'}\left[ \sup_{Z \in \mathcal{H}^{(l)}} \left| \epsilon'^\top Z \right| \right] + \|\mathcal{H}^{(l)}\|_2 \right)
$
for a universal constant $C$, where we used $2p \leq 1$.

Since the output is bounded by the assumption on the activation function, we have $\sup_{Z \in \mathcal{H}^{(l)}}\|Z\|_2 \leq \sqrt{N}R =: D$.  
Now substituting the above result back to \eqref{eq:reduction_rad-step1}, we have
\begin{equation} 
\begin{split}
& Q^{-1}\mathcal{R}(\mathcal{\tilde{H}}^{(l+1)}, p)   \\
&\leq B^{(l)} (1 - \alpha)   \bigg[C \sqrt{\mathrm{Tr}[\tilde{A}(\mu^{(l)})^{2}]}  \bigg( \mathbb{E}_{\boldsymbol{\epsilon}}\sup_{Z \in \mathcal{H}^{(l)}}  \bigg| \sum_{n=1}^{N} \epsilon_{n} Z_{n} \bigg| + D \bigg) \bigg] 
 + \alpha B^{(l)} \mathbb{E}_{\boldsymbol{\epsilon}}  \bigg[\sup_{H \in \mathcal{\tilde{H}}^{(0)} } \bigg| \sum_{n=1}^{N} \epsilon_{n}   H_n \bigg|\bigg]  \\
   &  \leq B^{(l)} (1 - \alpha) \bigg[ C\sqrt{ \sum_{i=1}^{N} \bigg(\sum_{k=0}^{K-1} \mu_{k}^{(l)} \lambda_{i}^{k}\bigg)^{2} }    \bigg(  \mathbb{E}_{\boldsymbol{\epsilon}} \sup_{Z \in \mathcal{H}^{(l)}} \bigg|   \sum_{n=1}^{N} \epsilon_{n} Z_{n}  \bigg| + D\bigg) \bigg] \\  &\qquad\qquad\qquad\qquad\qquad\qquad\qquad\qquad\qquad\qquad 
   + \alpha B^{(l)} \mathbb{E}_{\boldsymbol{\epsilon}}  \bigg[\sup_{H \in \mathcal{\tilde{H}}^{(0)} } \bigg| \sum_{n=1}^{N} \epsilon_{n}   H_n \bigg|\bigg] \\
& = CB^{(l)} (1 - \alpha) \left[ \bigg(\sum_{i=1}^{N} \sum_{k=0}^{K-1} \mu_{k}^{(l)}  |\lambda_{i}|^{k} \bigg) \big( Q^{-1}\mathcal{R}(\mathcal{H}^{(l)},p)  +  D \big) \right]
+  \alpha B^{(l)} \mathbb{E}_{\boldsymbol{\epsilon}}  \bigg[\sup_{H \in \mathcal{\tilde{H}}^{(0)} } \bigg| \sum_{n=1}^{N} \epsilon_{n}   H_n \bigg|\bigg],
\end{split} \label{eq:reduction_rad}
\end{equation}
where $\lambda_{t}$ is the $t$-th eigenvlaue of $\tilde{A}$ and we have used the used $\sqrt{ \sum_{i=1}^{N} (\sum_{k=0}^{K-1} \mu_{k}^{(l)} \lambda_{i}^{k})^{2}} \leq \sum_{i=1}^{N} \sqrt{ (\sum_{k=0}^{K-1} \mu_{k}^{(l)} \lambda_{i}^{k})^{2} } = \sum_{i=1}^{N} \big| \sum_{k=0}^{K-1} \mu_{k}^{(l)} \lambda_{i}^{k} \big| \leq \sum_{i=1}^{N} \sum_{k=0}^{K-1} \mu_{k}^{(l)}  |\lambda_{i}|^{k}$.

Since $\sigma$ is $1$-Lipschitz using the contraction property from Proposition 10 of \citep{OonoS20}, we have
\begin{equation*}
\mathcal{\bar{R}}(\mathcal{H}^{(l+1)},p) \leq 2\mathcal{\bar{R}}(\mathcal{\tilde{H}}^{(l+1)},p), 
\end{equation*} 
leading to reduction in \eqref{eq:reduction_rad} to
\begin{multline}
Q^{-1}\mathcal{\bar{R}}(\mathcal{\tilde{H}}^{(l+1)}, p) \leq C2 B^{(l)} (1 - \alpha) \left[ \bigg(\sum_{i=1}^{N} \sum_{k=0}^{K-1} \mu_{k}^{(l)}  |\lambda_{i}|^{k} \bigg) \Big( Q^{-1}\mathcal{\bar{R}}(\mathcal{\tilde{H}}^{(l)},p) + D \Big) \right]\\
+ \alpha B^{(l)} \mathbb{E}_{\boldsymbol{\epsilon}}  \bigg[\sup_{H \in \mathcal{\tilde{H}}^{(0)} } \bigg| \sum_{n=1}^{N} \epsilon_{n}  H_n \bigg|\bigg].
 \label{eq:rad_peeling}
\end{multline} 

Given that $\mathcal{F} = \mathcal{\tilde{H}}^{(L)}$, the final reduction using \eqref{eq:rad_peeling} leads to 
\begin{multline} 
Q^{-1}\mathcal{\bar{R}}(\mathcal{F},p) 
\leq  C^{'}\alpha\sum_{l=1}^{L} (1-\alpha)^{l} 2^{l} \prod_{j=0}^{l-1} \left\{B^{(L-j)}\bigg(\sum_{i=1}^{N} \sum_{k=0}^{K-1} \mu_{k}^{(L-j)}  |\lambda_{i}|^{k} \bigg)\right\}  \mathbb{E}_{\boldsymbol{\epsilon}}  \bigg[\sup_{H \in \mathcal{\tilde{H}}^{(0)} } \bigg| \sum_{n=1}^{N} \epsilon_{n}  H_n \bigg|\bigg]\\
+  C^{'}(1-\alpha)\sum_{l=1}^{L} (1-\alpha)^{l} 2^{l} \prod_{j=0}^{l} B^{(L-j)} \bigg(\sum_{i=1}^{N} \sum_{k=0}^{K-1} \mu_{k}^{(L-j)}  |\lambda_{i}|^{k} \bigg) D.
\label{eq:rad_reduce} 
\end{multline}

By construction of $\mathcal{F}$, we know that $\mathcal{\bar{R}}(\mathcal{\tilde{H}}^{(0)},p) = \mathbb{E}_{\boldsymbol{\epsilon}}  \bigg[\sup_{H \in \mathcal{\tilde{H}}^{(0)} } \bigg| \sum_{n=1}^{N} \epsilon_{n}  H_n \bigg|\bigg]$, and we have that 
\begin{align*} 
\mathcal{\bar{R}}(\mathcal{\tilde{H}}^{(0)},p)  & 
\leq 2 \mathcal{\bar{R}}(\mathcal{H}^{(0)},p) 
= 2\mathbb{E}_{\boldsymbol{\epsilon}} \Bigg[\sup_{w \in \mathbb{R}^{C_0}:\|w\|_1 \leq B^{(0)}} 
\Bigg|  \sum_{n=1}^{N} \sum_{c=1}^{C_0} \epsilon_{n} X_{nc}  w_c \Bigg| \Bigg]  \\ &
= 2 B^{(0)} \mathbb{E}_{\boldsymbol{\epsilon}} \Bigg[\max_{c \in [C_0]} \Bigg|  \sum_{n=1}^{N} \epsilon_{n} X_{nc} \Bigg| \Bigg] 
\leq 2 B^{(0)}\mathbb{E}_{\boldsymbol{\epsilon}} \Bigg[\Bigg\| \sum_{n=1}^{N} \epsilon_{n} X_{n\cdot}   \Bigg\|_{2} \Bigg] \\
& \leq 2 B^{(0)} \sqrt{ \mathbb{E}_{\boldsymbol{\epsilon}} \sum_{c=1}^{C_0} \Bigg( \sum_{n=1}^{N} \epsilon_{n} X_{nc}  \Bigg)^{2} }  \;\; \mathrm{(\because~Jensen\;Inequality)}\\
& = 2B^{(0)}\sqrt{ \mathbb{E}_{\boldsymbol{\epsilon}} \sum_{c=1}^{C_0} \sum_{n,m=1}^{N}  \epsilon_{n}\epsilon_{m} X_{nc}X_{mc}   } 
= 2B^{(0)}\sqrt{  \sum_{c=1}^{C_0} \sum_{m=1}^{N} 2p (X_{mc})^{2}   } \\
& = 2 B^{(0)}\sqrt{  2p   } \| X \|_{\mathrm{F}}. 
\end{align*}


Given that $p = p_0 = \frac{MU}{(M+U)^2}$, we have
\begin{align}
\mathcal{\bar{R}}(\mathcal{\tilde{H}}^{(0)},p_0) \leq  2 B^{(0)}\sqrt{\frac{2MU}{(M+U)^2}}\| X \|_{\mathrm{F}}. \label{eq:rad_h0}
\end{align}

By combining \eqref{eq:rad_h0} with \eqref{eq:rad_reduce}, the resulting final Rademacher complexity bound is given by
\begin{align} 
 Q^{-1} \mathcal{R}(\mathcal{F},p_0)  \notag  
\leq & Q^{-1}\mathcal{\bar{R}}(\mathcal{F},p_0) \notag \\
\leq & C^{'} \Bigg\{\sqrt{\frac{2MU}{(M+U)^2}}B^{(0)}\alpha\sum_{l=1}^{L} (1-\alpha)^{l} 2^{l} \prod_{j=0}^{l-1} \left\{ B^{(L-j)}\bigg(\sum_{i=1}^{N} \sum_{k=0}^{K-1} \mu_{k}^{(L-j)}  |\lambda_{i}|^{k} \bigg)\right\}  \| X\|_{\mathrm{F}} \notag \\
& +  (1-\alpha)\left[ \sum_{l=1}^{L} (1-\alpha)^{l} 2^{l} \prod_{j=0}^{l} B^{(L-j)} \bigg(\sum_{i=1}^{N} \sum_{k=0}^{K-1} \mu_{k}^{(L-j)}  |\lambda_{i}|^{k} \bigg)   \right]D\Bigg\}, \label{eq:gen_gcnII_gpr} 
\end{align}
by redefining the universal constant $C'$ if necessary.

}

GCNII
\textit{Proof of Corollary 1.}{
By replacing the generalized Pagerank $\tilde{A}(\mu)$ with the normalized adjacency matrix $\tilde{A}$, which is equivalent to setting $\mu_{1}^{(l)} = 1$ and rest of the elements in $\mu^{(l)}$ to zero, we obtain the desired result.
}

\section{Summary of Hyperparamter Selection}
In this section we discuss provide the details of hyperparameters selected for the proposed method though the validation process.

\begin{table}[b]
\centering
\begin{tabular}{l|l|l|l|l|l|l|l|l}
Dataset   & GPR Coeffs. & LR & $WD_1$ & $WD_2$ & $WD_3$ & $\lambda$ & $\alpha$ & Dropout         \\\hline
Cora & 4 & 0.01 & 1.0 & 0.0001  & 0.1 & 0.1 & 0.3 & 0.6       \\  
Citeseer & 16 & 0.01 & 1.0 & 0.0001 & 0.1 & 0.5   & 0.1     &  0.1       \\  
Pubmed & 3 & 0.01 & 0.0001 & 0.0001  & 0.1 & 0.1 & 0.1 &      0.5   \\  
\end{tabular} \caption{Hyperparameters for semi-supervised node-classification} \label{table:hyper_full}
\end{table}

\begin{table}[t]
\centering
\begin{tabular}{l|l|l|l|l|l|l|l}
Dataset   & GPR Coeffs. & layers & LR & Weight Decay & $\lambda$ & $\alpha$ & Dropout   \\\hline
Cora & 3 & 64 & 0.01 & 0.0001 & 0.5  & 0.1  & 0.5 \\
Citeseer & 2 & 64 & 0.01 & 0.0001  & 0.5 & 0.4     &  0.7    \\
Pubmed & 3 & 4 &  0.01   & 0.0001   & 0.5 & 0.5   &  0.2      \\
Chameleon & 3 & 2 & 0.01 & 0.001  & 1.5    & 0.6 &  0.6   \\
Cornell & 2 & 4 &0.01 & 0.0001   &  1.0  & 0.9 & 0.4    \\
Texas & 4 & 4 & 0.01 & 5e-4  & 1.0   & 0.5 &  0.5   \\
Wisconsin & 3 & 16 & 0.01 & 5e-5   & 1.5  & 0.6 & 0.3     \\
\end{tabular}
 \caption{Hyperparameters for fully-supervised node-classification} \label{table:hyper_semi}
\end{table}

For both experiments, we tuned the number of coefficients of the GPR as a hyperparamter selection from the set of $\{2,3,4,8,16,32\}$. For fully-supervised node-classification, we used the same parameter ranges as in GCNII \citep{pmlr-v119-chen20v}; $64$ hidden units, learning  rate $0.01$, the number of layers from $(2,4,8,16,32,64)$, $\lambda \in (0.5,1.0,1.5)$, $\alpha \in (0.1,0.2,\ldots,0.9)$, dropout $\in (0.1,0.2,\ldots,0.9)$, and weight decay $\in (0.001,5\mathrm{e-}3,\dots,1\mathrm{e-}6)$. 

For semi-supervised node-classification,  we fixed the  learning  rate with $0.01$ and $\alpha =0.1$ as as given in \citep{pmlr-v119-chen20v}. We set the weight decay rate $WD_2 = 0.0001$ and  applied hyperparameter tuning for weight decays for $WD_1$ and $WD_3$ from the set $(1.0,0.1,0.01,\ldots,0.0001)$. Further we performed hyperparameter tuning for $\lambda \in (0.1,0.2,\ldots,0.9)$, dropout $ \in (0.1,0.2,\ldots,0.9)$, and number of coefficients of the GPR $K$ from the set $(2,3,4,8,16)$.

Details of the parameters selected using hyperparamter tuning for semi-supervised node classification and fully-supervised node-classification by the AdaGPR are listed in  Tables \ref{table:hyper_semi} and \ref{table:hyper_full}, respectively.

\section{Further Analysis of Trained Models}
Table \ref{table:cora_coeffs} shows  coefficients of a semi-supervised learning model for Cora with 8 layers and 4 Pagerank coefficients. Though there are no sparseness among coefficients, notice that there is a gradual change of coefficients from shallow layers to deep layers.  As the layers increase from the first to the seventh layers the largest coefficient shifts  between the first two coefficients, while the forth coefficient gradually decreases. Recall that the coefficient at $0$ represent the identity matrix with no graph convolution, hence, indicates that each layer need not have graph convolution. 

\begin{table*}[t]
\centering
\begin{tabular}{l|cccc}\hline
\multirow{2}{*}{Layers} & \multicolumn{4}{l}{GPR Coeff.} \\ \cline{2-5} 
  & 0  & 1  & 2 & 3  \\ \hline
1 & 0.2664 &  0.2606 &  0.2449& 0.2279\\
2 & 0.2755 & 0.2601 & 0.2435 &  0.2207\\
3 & 0.2626 &  0.2733 & 0.2438 & 0.2201\\
4 & 0.2863 & 0.2574 & 0.2467 & 0.2093\\
5 & 0.2412 & 0.2861 & 0.2537 & 0.2188\\
6 & 0.2588 & 0.2726 & 0.2574 & 0.2111\\
7 & 0.2664 & 0.2854 & 0.2463 & 0.2017\\
8 & 0.1407 & 0.2933 & 0.2919 & 0.2740\\
\hline\end{tabular}
\caption{GPR coefficients of Cora}\label{table:cora_coeffs}
\end{table*}

Table \ref{table:cite_coeff} shows the GPR coefficients for semi-supervised node classification for Citeseer dataset using 16 payers and 16 GPR coefficients. Notice that  coefficients in shallow layers, layer 1 to layer 8, roughly equal to $1/16$. By analyzing the learning parameters for coefficients, we found that this is due to small values of the learning parameters in shallow layers. This may have caused by the application of softmax-like  (sparsemax) activation to a set of values that are close to zeros. As the layers increases beyond 8, coefficients start to deviate and it becomes clear that each layer applies a convolution with a different generalized Pagerank. Furthermore, it is worth noticing that with the increase in layers the value of the first coefficient becomes prominent and the coefficients for the higher order terms gradually decreases. 

\begin{sidewaystable}
\begin{tabular}{l|llllllllllllllll}
\multirow{2}{*}{Layers} & \multicolumn{7}{l}{GPR Coeff.} \\ \cline{2-17} 
              & 0   & 1  & 2  & 3 & 4 & 5 & 6 & 7 & 8  & 9  & 10 & 11 & 12 & 13 & 14 & 15    \\ \hline
1 &0.063 &0.063 &0.063 &0.063 &0.063 &0.063 &0.063 &0.062 &0.062 &0.062 & 0.062 &0.062 &0.062 &0.062 &0.062 &0.062 
\\
2 &0.063 & 0.063 &0.063 & 0.063 & 0.063 &0.063 & 0.063 & 0.062 &0.062 & 0.062 & 0.062 & 0.062 & 0.062 & 0.062 & 0.062 & 0.062 
\\
3 &0.063 & 0.063 & 0.063 & 0.063 & 0.063 & 0.063 & 0.063 &  0.062 & 0.062 & 0.062 & 0.062 & 0.062 & 0.062 & 0.062 & 0.062 & 0.062 
\\
4 &0.063 & 0.063 & 0.063 &  0.063 & 0.063 & 0.063 & 0.063 & 0.062 & 0.062 & 0.062 & 0.062 & 0.062 & 0.062 & 0.062 & 0.062 & 0.062 
\\
5 &0.064 & 0.064 & 0.063 &  0.063 & 0.063 & 0.063 & 0.063 &  0.063 & 0.062 & 0.062 & 0.062 & 0.062 & 0.062 & 0.062 & 0.062 & 0.061 
\\
6 &0.064 &0.064 &0.064 &0.064 &0.063 &0.063 &0.063 &0.062 &0.062 &0.062 &0.062 & 0.062 & 0.061 &  0.062 &  0.061 &  0.061 
\\
7 &0.065 &0.065 &0.064 &0.064 &0.064 &0.063 &0.063 &0.063 &0.062 &0.062 &0.062 &0.061 &0.061 &0.061 &0.060 &0.060  
\\
8 &0.066 &0.066 &0.066 &0.065 &0.064 &0.064 &0.063 &0.063 &0.062 &0.062 &0.061 &0.060 &0.060 &0.060 &0.060 &0.059 
\\
9 &0.068 &
0.069 &
0.067 &
0.066 &
0.065 &
0.064 &
0.063 &
0.063 &
0.062 &
0.061 &
0.061 &
0.060 &
0.059 &
0.059 &
0.058 &
0.056 
\\
10 &0.072 &
0.070 &
0.069 &
0.067 &
0.065 &
0.065 &
0.064 &
0.063 &
0.062 &
0.061 &
0.060 &
0.058 &
0.058 &
0.057 &
0.056 &
0.055 
\\
11 &0.077 &
0.075 &
0.072 &
0.069 &
0.070 &
0.067 &
0.064 &
0.062 &
0.060 &
0.059 &
0.057 &
0.056 &
0.055 &
0.054 &
0.052 &
0.052 
\\
12 &0.090 &
0.084 &
0.078 &
0.074 &
0.071 &
0.067 &
0.064 &
0.061 &
0.059 &
0.056 &
0.054 &
0.052 &
0.050 &
0.048 &
0.046 &
0.046 
\\
13 &0.135 &
0.102 &
0.091 &
0.080 &
0.074 &
0.068 &
0.062 &
0.057 &
0.055 &
0.049 &
0.045 &
0.042 &
0.040 &
0.036 &
0.033 &
0.030 
\\
14 &0.280 &
0.132 &
0.107 &
0.085 &
0.072 &
0.061 &
0.052 &
0.044 &
0.038 &
0.032 &
0.027 &
0.022 &
0.018 &
0.014 &
0.011 &
0.007 
\\
15 &0.557 &
0.147 &
0.109 &
0.069 &
0.050 &
0.032 &
0.021 &
0.010 &
0.001 &
0.001 &
0.002 &
0.001 &
0.000 &
0.000 &
0.001 &
0.000 
\\
16 &0.879 &
0.080 &
0.039 &
0.002 &
0.000 &
0.000 &
0.000 &
0.000 &
0.000 &
0.000 &
0.000 &
0.000 &
0.000 &
0.000 &
0.000 &
0.000 
\end{tabular}
\caption{GPR coefficients of a trained AdaGPR model for Semi-supervised learning with Citeseer.} \label{table:cite_coeff}
\end{sidewaystable}

Tables \ref{table:cham_coeffs} and \ref{table:texas_coeffs} further show  learned Pagerank coefficients of trained models for  Chameleon and Texas under fully-supervised node-classification. 
The trained model for Chameleon has  graph convolution only at the second layer with the normalized adjacency matrix and the first layer act as a  residual layer. The learned model for Cornell shows that only the first two layers apply  graph convolutions with gradual adaptations of the GPR from shallow layers to deeper layers. An interesting observation is  with the trained model for  Texas, where it has no graph convolution in all four layers. By looking at these sparse GPR coefficients one may draw a conclusion that many of the above models (e.g. Texas) do not need any graph convolution at all. We have found that graph convolutions with higher orders are important during the learning process though the final trained model may have zeros or small values. In Figures \ref{fig:texas_trains1},\ref{fig:texas_trains2},\ref{fig:texas_trains3},\ref{fig:texas_trains4}, we show the change of values in GPR coefficients at each iteration with fully-supervised node classification for Texas  with a AdaGPR model that consists of  4 convolution layers and 4 GPR coefficients.

\begin{table*}[!]
\centering
\begin{tabular}{l|ccc}\hline
\multirow{2}{*}{Layers} & \multicolumn{3}{l}{GPR Coeff.} \\ \cline{2-4} 
  & 0  & 1  & 2  \\ \hline
1 & 1 &0 & 0\\
2 & 0 & 1 & 0\\
\hline\end{tabular}
\caption{GPR coefficients of Chameleon}\label{table:cham_coeffs}
\end{table*}

\begin{table*}[!]
\centering
\begin{tabular}{l|cccc}\hline
\multirow{2}{*}{Layers} & \multicolumn{4}{l}{GPR Coeff.} \\ \cline{2-5} 
  & 0  & 1  & 2  & 3 \\ \hline
1 & 1 &0 & 0& 0\\
2 & 1 & 0 & 0& 0\\
3 & 1 &0 & 0& 0\\
4 & 1 & 0 & 0& 0\\
\hline\end{tabular}
\caption{GPR coefficients of Texas}\label{table:texas_coeffs}
\end{table*}

\begin{subfigures}
\begin{figure}
\centering
{\includegraphics[width = 3.5in]{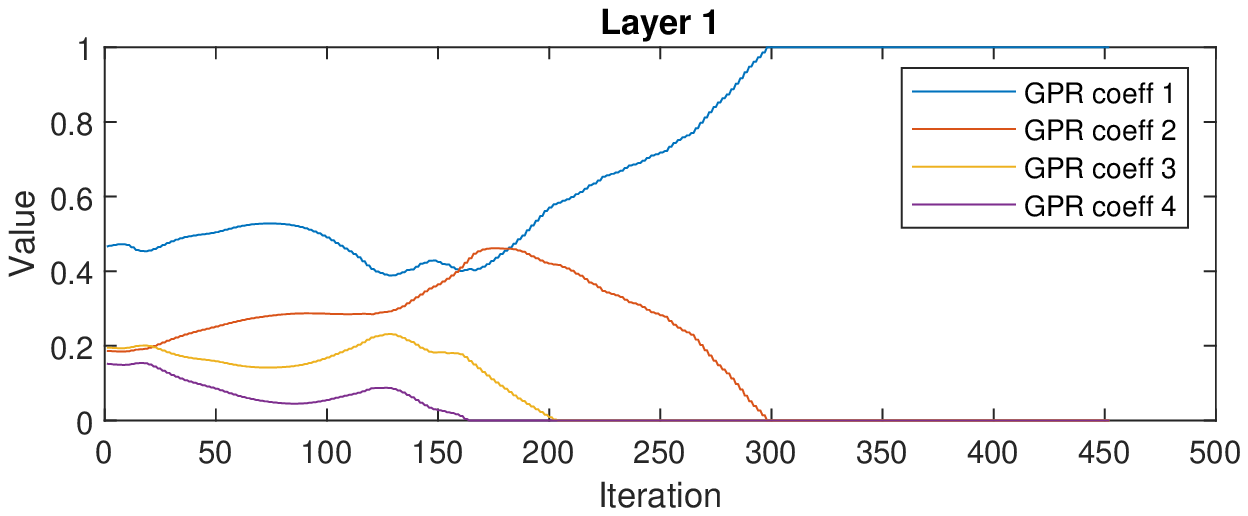}}
\caption{Coefficient evolution of Layer 1 for Texas}
\label{fig:texas_trains1}
\end{figure}
\begin{figure}
\centering
{\includegraphics[width = 3.5in]{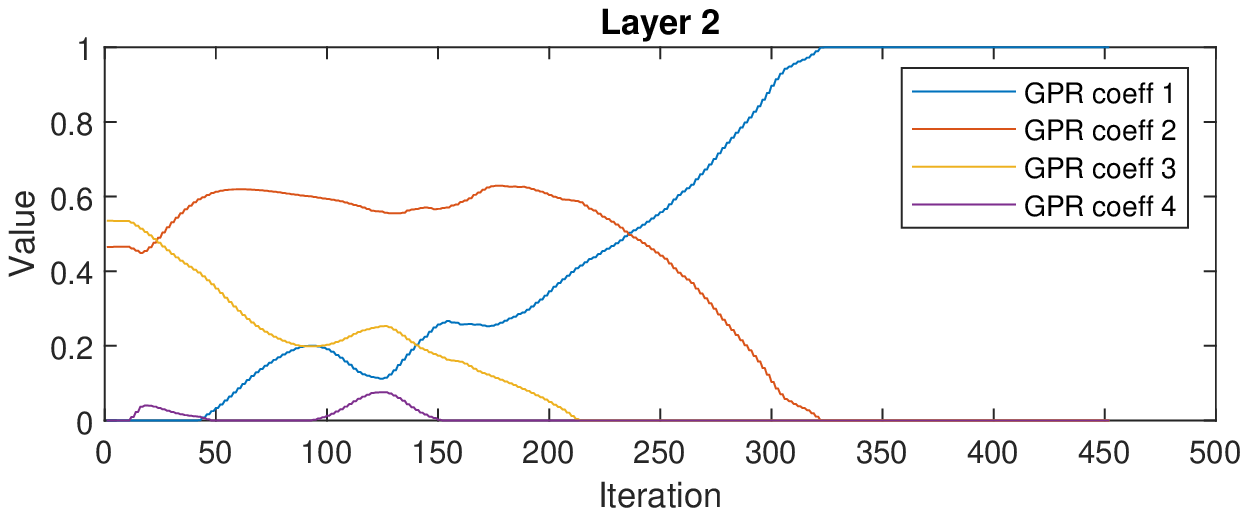}}
\caption{Coefficient evolution of Layer 2 for Texas}
\label{fig:texas_trains2}
\end{figure}
\begin{figure}
\centering
{\includegraphics[width = 3.5in]{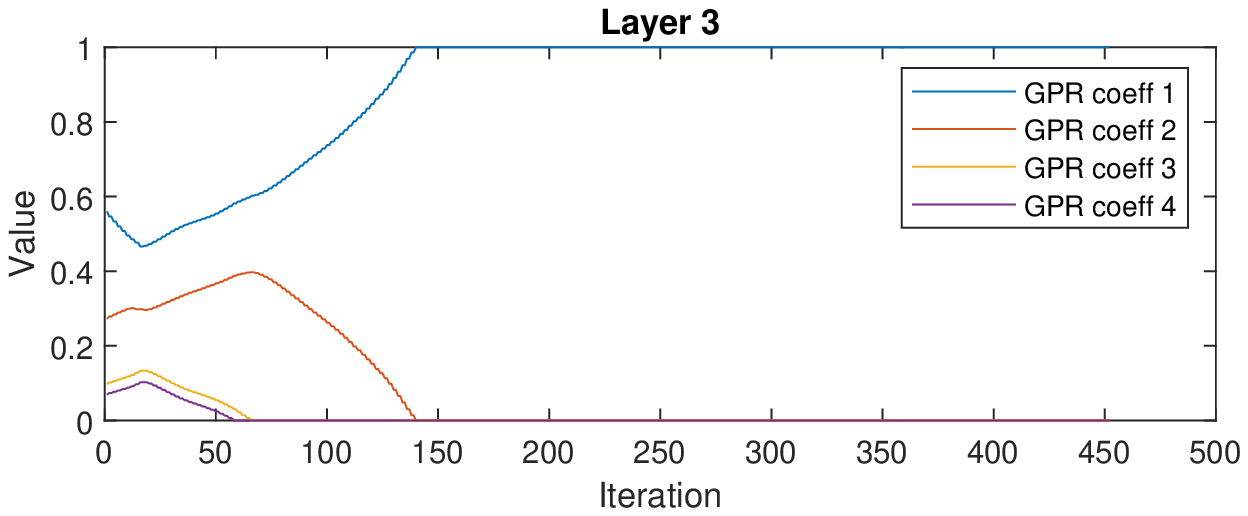}}
\caption{Coefficient evolution of Layer 3 for Texas}
\label{fig:texas_trains3}
\end{figure}
\begin{figure}
\centering
{\includegraphics[width = 3.5in]{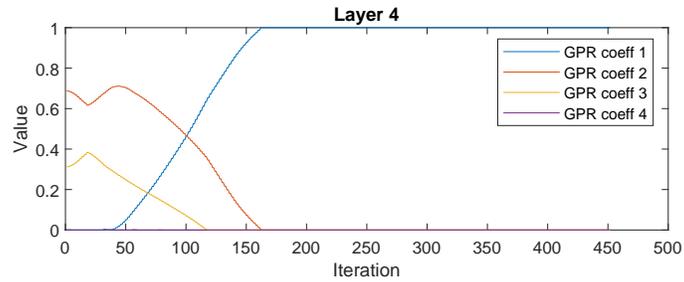}}
\caption{Coefficient evolution of Layer 4 for Texas}
\label{fig:texas_trains4}
\end{figure}
\end{subfigures}

\section{Ablation Studies}

We conducted ablations studies to understand the oversmoothing effect under layer-wise adaptive learning of AdaGPR. We compared AdaGPR with vanilla GCN and GPR convolution without adapted layer-wise coefficients. It is difficult to design a general GPR convolution with appropriate user specified coefficients. For simplicity, we considered the spacial case where all GPR coefficients are equal and assigned values of $1/\mathrm{(number\;of\;GPR\;coefficients)}$. 

Figure \ref{fig:ablation1},\ref{fig:ablation2} shows ablation plots of Cora and Citeseer for semi-supervised node classification. We can see that adaptive learning with AdaGPR improves accuracy with the increase of layers. Adaptive layer-wise learning of GPR coefficients consistently improve accuracy with the increasing number of layers compared to having constant GPR coefficients.

\begin{subfigures}[b]
\begin{figure}
\centering
{\includegraphics[width = 3.1in]{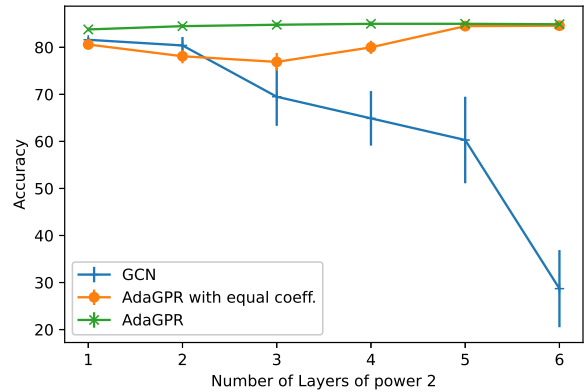}}
\caption{Ablation study of Cora}
\label{fig:ablation1}
\end{figure}
\begin{figure}
\centering
{\includegraphics[width = 3.1in]{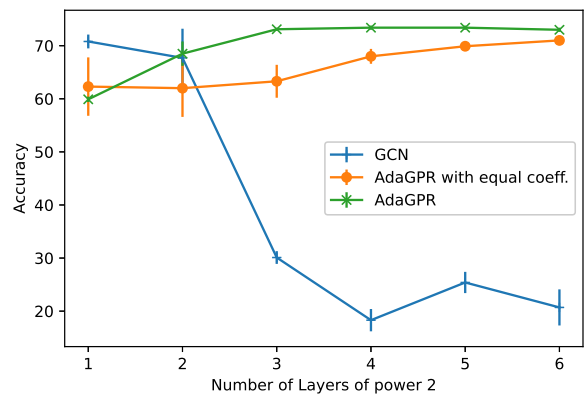}}
\caption{Ablation study of Citeseer}
\label{fig:ablation2}
\end{figure}
\end{subfigures}

\end{document}